\documentclass{article}
\usepackage{spconf,amsmath,graphicx,hyperref}
\usepackage{cite}
\usepackage{amsmath,amssymb,amsfonts}
\usepackage{graphicx}
\usepackage{textcomp}
\usepackage{amsmath}
\usepackage{booktabs} 
\usepackage{multirow}
\usepackage{subcaption}
\usepackage{hyperref}
\usepackage{tabularx}
\usepackage{url}            
\usepackage{booktabs}       
\usepackage{amsfonts}       
\usepackage{nicefrac}       
\usepackage{microtype}      
\usepackage{xcolor}         
\usepackage{algorithm}
\usepackage{colortbl}
\usepackage{algpseudocode}
\usepackage{amsmath}
\usepackage{tikz}
\usepackage{orcidlink}
\usepackage[font=small]{caption}
\usepackage{float}
\usepackage{url}
\usepackage{stfloats}

\usepackage{orcidlink}
\usepackage{comment}
\usepackage{amssymb}
\usepackage{graphicx}
\usepackage{braket}
\usepackage[table]{xcolor}
\usepackage[skip=0pt,font=small]{caption}
\usepackage{booktabs}
\usepackage{setspace}
\usepackage{wrapfig}
\usepackage{tabularx}
\usepackage{xcolor}

\title{AERIAL: Adversarial Evaluation of Robustness in Accuracy-Preserving Low-Precision EEG Decoders}

\name{
Saim Rehman, 
Muhammad Shafique
}

\address{
\small eBRAIN Lab, Division of Engineering, New York University Abu Dhabi (NYUAD), Abu Dhabi, UAE\\
\small \{sr7849, muhammad.shafique\}@nyu.edu
}

\begin{document}
%
\maketitle

\begin{abstract}
Deployment-oriented compression is attractive for resource-constrained
brain--computer interfaces (BCIs), but whether it changes adversarial
vulnerability remains unclear. On BCI Competition IV-2a, we compare
32-bit floating-point (FP32) EEGNet and ShallowConvNet models with global
magnitude pruning and simulated INT8 post-training quantization (PTQ) and
quantization-aware training (QAT) across nine subjects and three seeds.
Simulation provides differentiable quantize--dequantize models for
white-box attacks and gradient analysis, while native TensorRT deployment
is used for validation. Accuracy-preserving compression does not improve
direct robustness: at $\epsilon=0.005$, EEGNet PGD accuracy remains
22--24\% across FP32, 50\% pruning (P50), PTQ, and QAT. However, P50
reduces bidirectional transfer efficiency to 0.963/0.928
(FP32$\rightarrow$P50/P50$\rightarrow$FP32), versus 0.994/0.997 for PTQ;
the same trend holds for ShallowConvNet. Gradient alignment shows a
corresponding separation, while native PTQ agrees with simulated
clean/adversarial predictions in 95--98\% of cases. These results show
that direct robustness, adversarial transfer, and deployment efficiency
are distinct properties of compressed EEG decoders.
\end{abstract}
\begin{keywords}
EEG, Brain--Computer Interface, Adversarial Robustness, Model Pruning, Quantization.
\end{keywords}
\section{Introduction}
Motor-imagery EEG decoders increasingly rely on compact convolutional networks such as EEGNet and ShallowConvNet \cite{tangermann2012,lawhern2018,schirrmeister2017}. Yet EEG classifiers are vulnerable to small adversarial perturbations and transferable attacks \cite{zhang2019}. Recent work has therefore emphasized adversarial training, detection, and benchmark design for secure BCIs \cite{chen2024abat,chendefense2025}. In parallel, deployment on low-power hardware motivates pruning and low-precision inference \cite{han2015,jacob2018,nagel2021}. These two concerns intersect: pruning or quantization may alter gradients and attack transfer even when task accuracy is unchanged, and quantized models can exhibit misleading apparent robustness under weak gradient evaluations \cite{gupta2022,bernhard2019,piras2025}.

Recent EEG adversarial studies have primarily focused on improving
robustness through adversarial training, defense benchmarking, or
robust architectures \cite{chen2024abat,chendefense2025,samuel2026hcnn},
while edge-oriented EEG work has studied quantization and model
compression mainly from the perspective of accuracy, memory, latency,
and energy \cite{schneider2020qeegnet,wang2024mibminet}. These two lines
of work leave a distinct methodological gap: it remains unclear whether
\emph{accuracy-preserving compression} changes direct white-box
vulnerability and cross-model adversarial transfer under otherwise
matched EEG decoders, and whether conclusions drawn from differentiable
quantization simulation persist under a native edge runtime. We address
this intersection by holding the data protocol, decoder family, and
attack configuration fixed while varying compression mechanism, and by
separately validating simulated PTQ with native TensorRT execution.
Accordingly, we ask: \emph{for EEG motor-imagery decoders, when a
compression method preserves clean accuracy, does it also change the
adversarial attack surface?}

\textbf{Our novel contributions, henceforth, are:} 
\begin{itemize}
    \item Providing a systematic study of how neural-network compression alters the adversarial attack surface of motor-imagery EEG decoders on BCI Competition IV-2a, explicitly separating \emph{direct white-box robustness} from \emph{cross-model adversarial transferability}
    \item  To show that compression methods with nearly identical clean accuracy and direct adversarial robustness can exhibit substantially different transfer behavior: 50\% magnitude pruning consistently weakens adversarial transfer across distinct CNN decoder architectures, whereas simulated INT8 PTQ and QAT largely preserve FP32 transferability, which demonstrates that the observed compression-dependent separation is not specific to a single model architecture.
    \item Validating the simulated PTQ findings under native TensorRT
deployment on a Jetson Orin Nano. Native and simulated PTQ agree on
95--98\% of clean/adversarial predictions, while measured latency and
power reveal architecture-dependent INT8 deployment benefits.
    \item Investigating the mechanism underlying these differences through FP32--compressed input-gradient cosine similarity and sign agreement, showing that stronger pruning substantially disrupts local gradient alignment, while PTQ/QAT retain gradient geometry close to the FP32 models. This links compression-induced changes in adversarial transferability to changes in local input-gradient structure
    
\end{itemize}

\section{Experimental Design}
\subsection{Data, models, and compression}

We use BCI Competition IV-2a \cite{tangermann2012}: nine subjects,
four motor-imagery classes, 22 EEG channels at 250 Hz, and separate
training (T) and evaluation (E) sessions. Signals are band-pass filtered
to 4--38 Hz, converted to $\mu$V, exponentially standardized
($10^{-3}$; 1000-sample initialization), and segmented from $-0.5$ to
$4.0$ s relative to the cue. T is stratified 80/20 for train/validation
and E is used only for testing. We evaluate EEGNet \cite{lawhern2018}
and ShallowConvNet \cite{schirrmeister2017} over seeds
$\{1,48,550\}$.

We apply global unstructured $L_1$ magnitude pruning at 30/50/70\%
with recovery fine-tuning, and symmetric simulated INT8 Q/DQ using PTQ
and QAT. PTQ activation ranges use training data only; QAT fine-tunes the
calibrated model with straight-through gradients. P30/P50 preserve
EEGNet clean accuracy, whereas P70 collapses to 25.96\% and is treated
as over-compression rather than evidence of robustness.

\begin{figure*}[t]
    \centering
    \includegraphics[width=\linewidth]{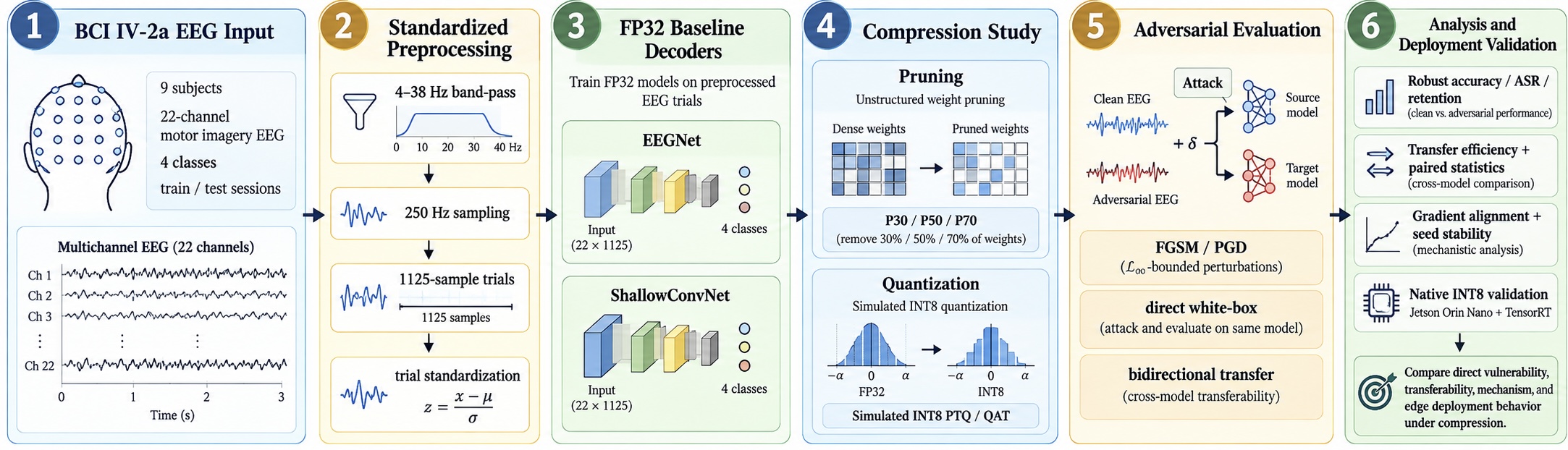}
    \caption{Full Methodology uncovering AERIAL's evaluation pipeline}
    \label{fig:meth}
\end{figure*}

\subsection{Threat model and analysis}
Attacks perturb the standardized digital decoder input under $L_\infty$ budgets $\epsilon\in\{.001,.0025,.005,.01\}$.
EEGNet uses FGSM and PGD; Shallow provides confirmatory PGD results at
$\{.001,.005,.01\}$. PGD uses 20 steps, five random restarts, random
initialization, and step size $\epsilon/4$ \cite{madry2018,carlini2019}.
Because attacks act after standardization, $\epsilon$ is dimensionless. We report clean/robust accuracy, ASR, and transfer efficiency
\[
\mathrm{TE} =
\frac{A_{\rm clean}-A_{\rm transfer}}
     {A_{\rm clean}-A_{\rm direct}},
\]
where TE$=1$ denotes transfer as effective as direct PGD. Gradient
cosine/sign agreement are computed on common-correct samples.
Seeds are averaged within subject before paired inference ($n=9$);
we use Wilcoxon tests with Holm correction, bootstrap CIs, and
rank-biserial effect size.

\section{Results}
\subsection{Compression preserves clean accuracy, not white-box robustness}

Table~\ref{tab:direct} separates accuracy-preserving compression from
over-compression. For EEGNet, P30, P50, PTQ, and QAT retain clean
accuracy within 0.99 pp of FP32, while direct PGD accuracy at
$\epsilon=0.005$ remains 22--24\%. In contrast, P70 collapses clean
accuracy to 25.96\% and is therefore treated as an over-compressed
operating point rather than evidence of robustness. ShallowConvNet shows
the same accuracy-preserving pattern for P50 and PTQ, with direct PGD
accuracy remaining near 42\%. Across $\epsilon$, the plotted accuracy-preserving variants closely track
FP32 (Fig.~\ref{fig:results}(a), indicating little compression-dependent
change in direct white-box vulnerability; the separation instead emerges
under cross-model transfer (Sec.~\ref{sec: rs}). Simulated PTQ/QAT provide differentiable
low-precision models for attack and gradient analysis, while PTQ is
validated separately under native TensorRT execution
(Sec.~\ref{sec:jetson}).

\begin{table}[t]
\caption{Clean and direct PGD accuracy (\%) at $\epsilon=0.005$.
P70 represents over-compression rather than a robust operating point.}
\label{tab:direct}
\centering
\footnotesize
\setlength{\tabcolsep}{4pt}
\renewcommand{\arraystretch}{0.90}
\begin{tabular}{llcc}
\toprule
Arch. & Variant & Clean & PGD \\
\midrule
EEGNet & FP32 & 57.39 & 22.83 \\
       & P30  & 57.47 & 23.06 \\
       & P50  & 58.38 & 23.60 \\
       & P70  & 25.96 & 24.01 \\
       & PTQ8 & 57.42 & 22.57 \\
       & QAT8 & 57.45 & 22.52 \\
\midrule
Shallow & FP32 & 57.73 & 42.39 \\
        & P50  & 57.47 & 42.12 \\
        & PTQ8 & 57.60 & 42.08 \\
\bottomrule
\end{tabular}
\end{table}

Figure~\ref{fig:meth} summarizes the complete evaluation pipeline from
preprocessing and FP32 training through compression, adversarial analysis,
and native deployment validation.

\subsection{Pruning changes transfer geometry and local gradient alignment}
\label{sec: rs}
\begin{figure*}[t]
    \centering
    \includegraphics[width=\linewidth]{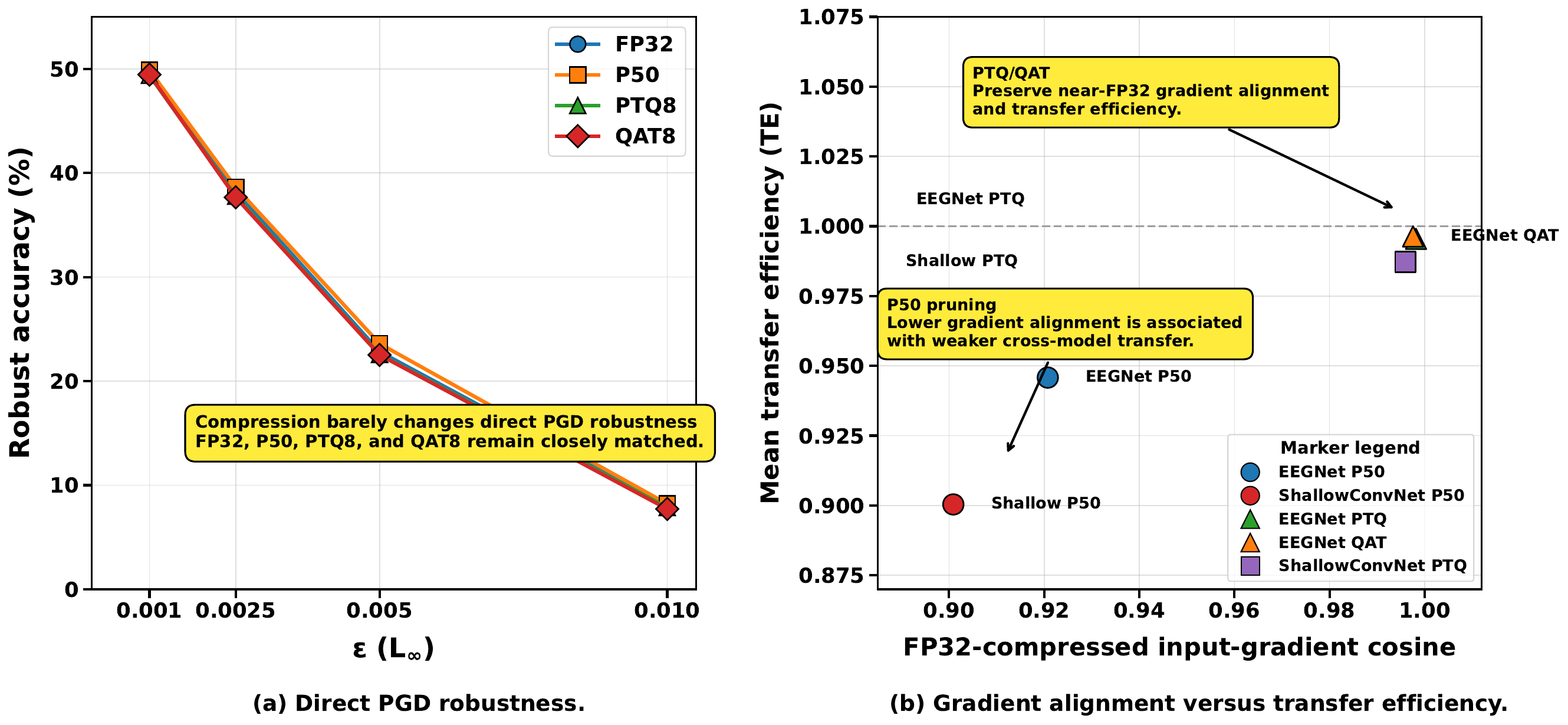}
    \caption{Direct vulnerability and compression-induced transfer geometry.}
    \label{fig:results}
\end{figure*}

The direct curves conceal a reproducible change in attack alignment. At $\epsilon=0.005$, EEGNet P50 transfers with TE 0.963 (FP32$\rightarrow$P50) and 0.928 (P50$\rightarrow$FP32), versus PTQ 0.994/0.997 and QAT 0.995/0.997. Shallow confirms the contrast: P50 TE is 0.897/0.904, versus PTQ 0.979/0.996. P50 gaps are positive for all nine subjects in both directions on both architectures (rank-biserial $=1$); the corresponding paired effects and
Holm-adjusted tests are reported in Table~\ref{tab:teall}.

Input-gradient diagnostics provide a mechanistic correlate. On a common-correct clean set, FP32--P50 cosine is 0.921 on EEGNet and 0.901 on Shallow, versus 0.998/0.996 for PTQ; EEGNet QAT is 0.998 and P30 is 0.990. Sign agreement follows the same pattern (P50 0.861/0.848 vs. PTQ 0.986/0.978). P50 cosine is lower than PTQ for all 9 subjects on both architectures (Holm $p=0.0156$), and lower than QAT on EEGNet by 0.0768. Figure~\ref{fig:results} shows that compression-level gradient alignment co-varies with transfer efficiency. Within-condition subject-wise Spearman correlations are not consistently significant, so we treat this as a local mechanistic probe rather than causal proof.

\subsection{Cross-epsilon consistency and attack-strength checks}
Table~\ref{tab:teall} shows that the transfer effect persists across all
three PGD budgets shared by the architectures and reports paired
$\Delta$pp and Holm-adjusted $p$ at $\epsilon=0.005$. P50 remains below
PTQ in every direction and budget. EEGNet P50 transfer is weakest for
P50$\rightarrow$FP32 at small-to-moderate budgets, while ShallowConvNet
shows a broader reduction (TE $\approx0.88$--$0.92$); PTQ remains near unity throughout. As attacks approach saturation, TE moves toward one as both direct and transferred attacks approach the accuracy floor.

\begin{table}[t]
\caption{Transfer efficiency (TE) across shared PGD budgets.
$\Delta$ and $p_H$ are reported at $\epsilon=0.005$;
$\Delta>0$ denotes weaker transfer than direct PGD.}
\label{tab:teall}
\centering
\scriptsize
\setlength{\tabcolsep}{1.7pt}
\renewcommand{\arraystretch}{0.90}
\begin{tabular}{llccccc}
\toprule
Arch. & Direction &
$.001$ & $.005$ & $.01$ &
$\Delta$ pp & $p_H$ \\
\midrule
EEGNet
& FP32$\rightarrow$P50 & .921 & .963 & .980 & 1.22 & .0156 \\
& P50$\rightarrow$FP32 & .897 & .928 & .964 & 2.47 & .0156 \\
& FP32$\rightarrow$PTQ & .981 & .994 & .999 & 0.21 & .0313 \\
& PTQ$\rightarrow$FP32 & 1.004 & .997 & 1.000 & 0.09 & .0938 \\
\midrule
Shallow
& FP32$\rightarrow$P50 & .914 & .897 & .918 & 1.57 & .0156 \\
& P50$\rightarrow$FP32 & .882 & .904 & .912 & 1.48 & .0156 \\
& FP32$\rightarrow$PTQ & .968 & .979 & .994 & 0.32 & .0156 \\
& PTQ$\rightarrow$FP32 & .992 & .996 & .999 & 0.06 & .0625 \\
\bottomrule
\end{tabular}
\end{table}

Attack-strength checks support the PGD configuration. In the EEGNet pilot at $\epsilon=0.005$, 10-, 20-, and 50-step PGD converge to the same mean robust accuracy (37.36\%), motivating 20 steps. On the full test set, PGD is at least as strong as FGSM at all four budgets; the FP32 robust-accuracy gap grows from 0.06 pp at $\epsilon=.001$ to 0.93 pp at $.01$. These checks reduce the risk that the compression comparison is an under-optimized first-order attack artifact.

\subsection{Quantization and gradient-masking check}
Quantized models do not exhibit the classic pattern of false robustness from unusable gradients \cite{athalye2018,gupta2022}. At EEGNet $\epsilon=.005$, direct PTQ/QAT robust accuracy is 22.57/22.52\%, FP32-crafted transfer gives 22.78/22.70\%, and reverse transfer to FP32 gives 22.92/22.89\% versus 22.83\% direct. QAT transfer efficiency is 0.995/0.997, and its FP32 gradient cosine is 0.998. Cross-precision attacks therefore remain nearly as effective as direct attacks, arguing that simulated INT8 preserves vulnerability rather than creating severe gradient obfuscation.

\subsection{Native edge-runtime validation}
\label{sec:jetson}

\begin{table}[t]
\caption{Native TensorRT PTQ fidelity. Accuracy and prediction agreement
are means over nine subjects using seed 1.}
\label{tab:jetson}
\centering
\footnotesize
\setlength{\tabcolsep}{2.5pt}
\renewcommand{\arraystretch}{0.92}
\begin{tabular}{lcc}
\toprule
Metric & EEGNet & Shallow \\
\midrule
Clean acc., sim./native (\%) & 59.22 / 58.99 & 58.37 / 57.45 \\
PGD acc., sim./native (\%)   & 27.58 / 27.78 & 42.36 / 42.90 \\
Clean agreement (\%)         & 97.42 & 94.91 \\
PGD agreement (\%)           & 97.84 & 94.98 \\
\bottomrule
\end{tabular}
\end{table}

To test whether the simulated PTQ conclusions persist under a real
deployment backend, we exported the seed-1 models to TensorRT 10.16 on a
Jetson Orin Nano operating in MAXN-SUPER mode. TensorRT was configured to
prefer INT8 execution with FP32 fallback where required, so this should
be interpreted as an INT8-preferred mixed-precision deployment rather
than a fully integer implementation. Clean and adversarial agreement were
evaluated over all nine held-out subjects, while latency and power were
measured on the representative A01 seed-1 model at batch size one. This
native experiment is therefore a deployment-fidelity check, not part of
the three-seed statistical robustness comparison.

Table~\ref{tab:jetson} shows that native execution closely preserves
simulated PTQ behavior: native-minus-simulated clean/PGD accuracy differs
by only $-0.23/+0.19$ pp for EEGNet and $-0.93/+0.54$ pp for
ShallowConvNet, with clean/adversarial prediction agreement of
97.42/97.84\% and 94.91/94.98\%, respectively. Thus, the simulated
adversarial behavior largely survives the native backend.
\begin{figure}[t]
    \centering
\includegraphics[width=\columnwidth]{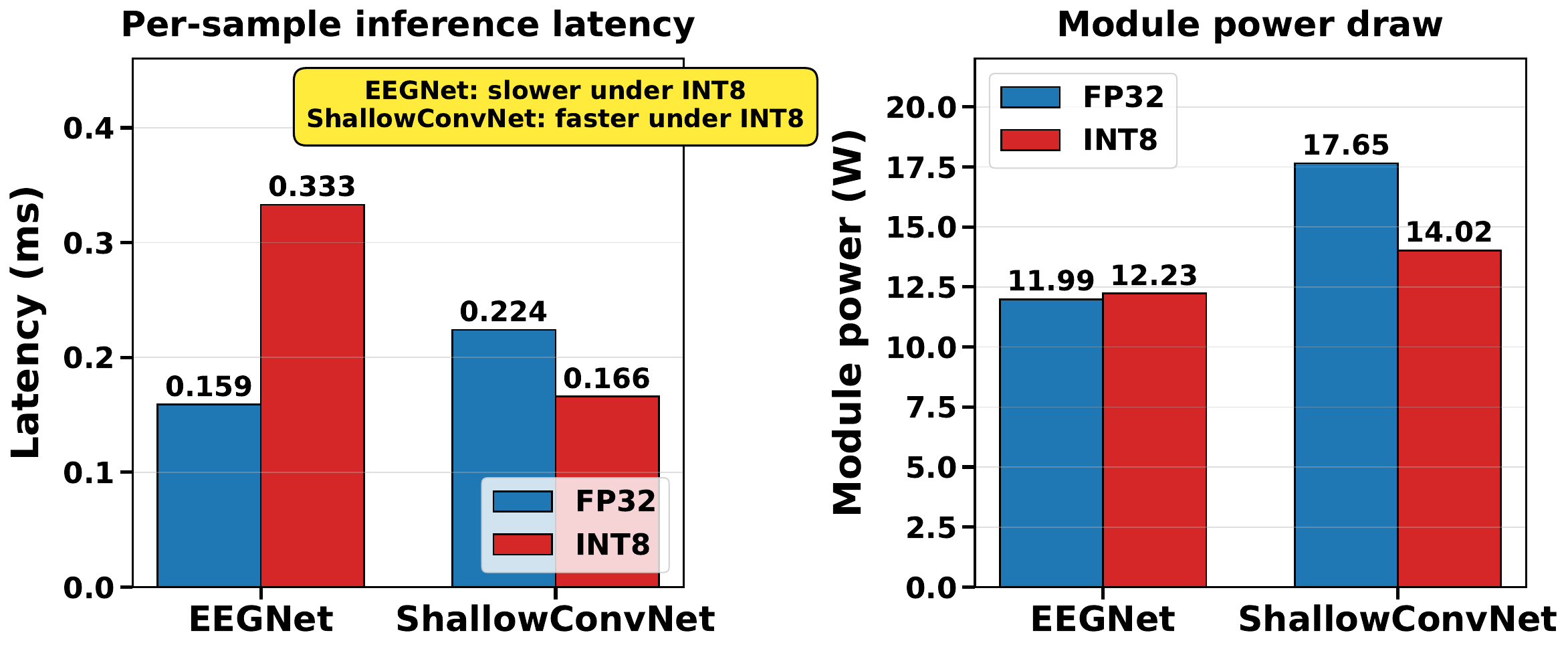}
    \caption{Native TensorRT runtime characteristics on Jetson Orin Nano
    at batch size one (A01, seed 1). INT8 denotes INT8-preferred
    mixed-precision execution with FP32 fallback.}
    \label{fig:hardware}
\end{figure}

Hardware efficiency, however, is architecture dependent
(Fig.~\ref{fig:hardware}). INT8 increases EEGNet latency by $2.09\times$
with essentially unchanged module power ($+2\%$), whereas ShallowConvNet
reduces latency by 26\% and module power by 21\%. Engine inspection
provides a plausible explanation: ShallowConvNet's main compute path
executes in INT8, while EEGNet retains FP32 fallback in its spatial
convolution and incurs additional reformatting. Numerical fidelity under
quantization therefore does not imply a uniform hardware benefit.

\section{Conclusion}
The results distinguish direct vulnerability from cross-model alignment.
P50 preserves clean and direct PGD accuracy yet consistently weakens
bidirectional transfer and FP32--compressed gradient alignment, whereas
PTQ/QAT remain close to FP32 in both properties. P30 causes little
separation, while P70 collapses clean accuracy near chance and is therefore
an over-compression failure rather than a robust operating point. Native
TensorRT results further show that simulated PTQ accurately predicts
adversarial behavior, but hardware efficiency depends on architecture.

Our scope is limited to BCI IV-2a motor imagery, two CNNs, global
unstructured magnitude pruning, symmetric INT8 quantization, and digital
first-order attacks on standardized inputs. Native validation covers PTQ
with one seed across nine subjects, with latency/power measured on A01;
we do not claim generality to structured pruning, other EEG paradigms,
physical attacks, quantization schemes, or accelerators. Overall,
accuracy-preserving compression is not an adversarial defense: pruning
can change transferable attack geometry without removing white-box
vulnerability, while INT8 largely preserves that geometry. Robustness,
transferability, and deployment efficiency should therefore be evaluated
as separate properties of compressed EEG decoders.

\section*{Acknowledgment}
 This work was supported in part by the NYUAD Center for CyberSecurity (CCS), funded by Tamkeen under the NYUAD Research Institute grant G1104. This research was carried out on the High Performance Computing resources at New York University Abu Dhabi.

\section*{Generative AI Use Disclosure}

During the preparation of this work, the authors used Generative AI tools (specifically ChatGPT and Grammarly) for language editing, text refinement, and visual refinement of the methodology figure. The authors reviewed and edited all generated or refined content as needed and take full responsibility for the publication’s content.

\vfill\pagebreak

\label{sec:refs}

\bibliographystyle{IEEEbib}
\bibliography{strings,refs}

\end{document}